\documentclass[letterpaper, 10 pt, conference]{ieeeconf}

\IEEEoverridecommandlockouts                              % This command is only needed if 
\title{\LARGE \bf
Teleoperated Robot Grasping in Virtual Reality Spaces
}

\usepackage[ruled,vlined]{algorithm2e}
\usepackage{graphicx}
\usepackage{xcolor}
\usepackage{url}

\author{Jiaheng Hu$^{1}$, David Watkins$^{1}$, and Peter Allen$^{1}$% <-this % stops a space
\thanks{$^{1}$Computer Science Department, Columbia University, New York, NY
    10027, USA 
        {\tt\small \{jh3916,djw2146\}@columbia.edu, allen@cs.columbia.edu}}%
}

\begin{document}

\maketitle
\thispagestyle{empty}
\pagestyle{empty}

%%%%%%%%%%%%%%%%%%%%%%%%%%%%%%%%%%%%%%%%%%%%%%%%%%%%%%%%%%%%%%%%%%%%%%%%%%%%%%%%
\begin{abstract}
Despite recent advancement in virtual reality technology, teleoperating a high DoF robot to complete dexterous tasks in cluttered scenes remains difficult. In this work, we propose a system that allows the user to teleoperate a Fetch robot to perform grasping in an easy and intuitive way, through exploiting the rich environment information provided by the virtual reality space. Our system has the benefit of easy transferability to different robots and different tasks, and can be used without any expert knowledge. We tested the system on a real fetch robot, and a video demonstrating the effectiveness of our system can be seen at \url{https://youtu.be/1-xW2Bx_Cms}
\end{abstract}

%%%%%%%%%%%%%%%%%%%%%%%%%%%%%%%%%%%%%%%%%%%%%%%%%%%%%%%%%%%%%%%%%%%%%%%%%%%%%%%%
\section{INTRODUCTION}
Robot teleoperation is a useful tool for handling tasks that are hard to be completed autonomously by robots, such as the DARPA Robotics Challenge \cite{Dellin-2014-7882}. Among different teleoperation methods, virtual reality based methods stand out as they provide an intuitive way for the user to examine the state of the robot and send commands \cite{surg}\cite{rosreality}\cite{lipton2017baxters}\cite{pub7803}. In spite of this, teleoperating robots with high degree of freedom to complete dexterous tasks such as grasping remains difficult, mainly due to the following reasons: 1. The high DoF of the robot makes it hard to create 1-1 mapping from the controller inputs to robot actions. 2. Different robots have different structures, making it difficult to switch between them. 3. Even given a 1-1 mapping between the controller inputs and robot actions, teleoperating a high DoF robot in cluttered environment would still be hard as it is usually unrealistic for the user to precisely predict the forward dynamics of the robot.

In this work, we extend upon Ros Reality \cite{rosreality} and create a robot teleoperation system that allows the user to easily and intuitively control a 13 DoF fetch robot to grasp objects in an unstructured environment. Instead of mapping controller inputs to robot actions, we only convey the user's intention of which object to grasp to the robot through utilizing the rich information provided by the virtual reality space. Thus, the user only interacts with the environment, and the robot plans the low-level controlling commands automatically, making our system easily extendable to different robots, and intelligent enough to handle dexterous grasping tasks. 

\section{Methods}

\begin{figure}[t]
\centering
\includegraphics[width=0.8\columnwidth]{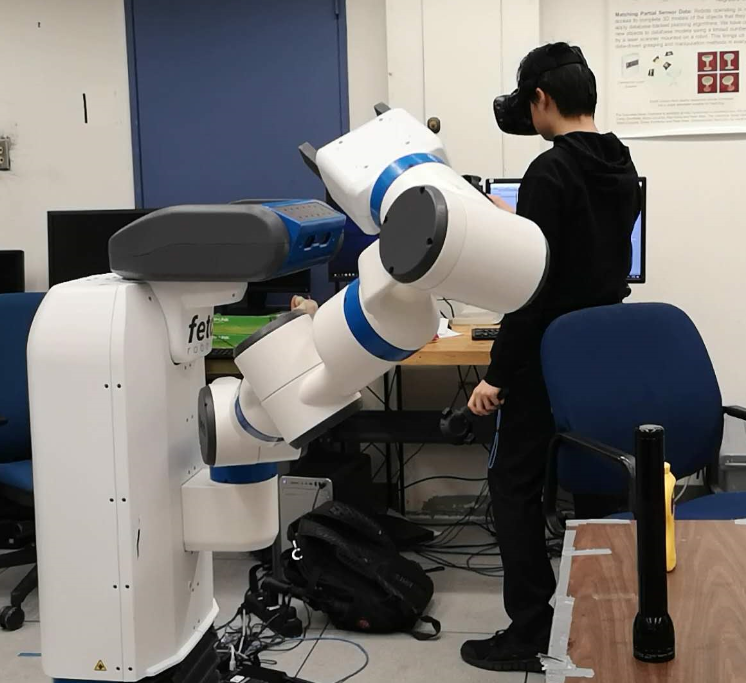} 
\caption{The Hardware Setup: the user is holding two vive controllers and wearing a vive headset, and the robot is facing a table with objects to grasp}
\label{fig:robot_and_user}
\end{figure}

\begin{figure}[t] 
  
  \begin{minipage}[b]{0.5\linewidth}
    \centering
    \includegraphics[width=.9\linewidth]{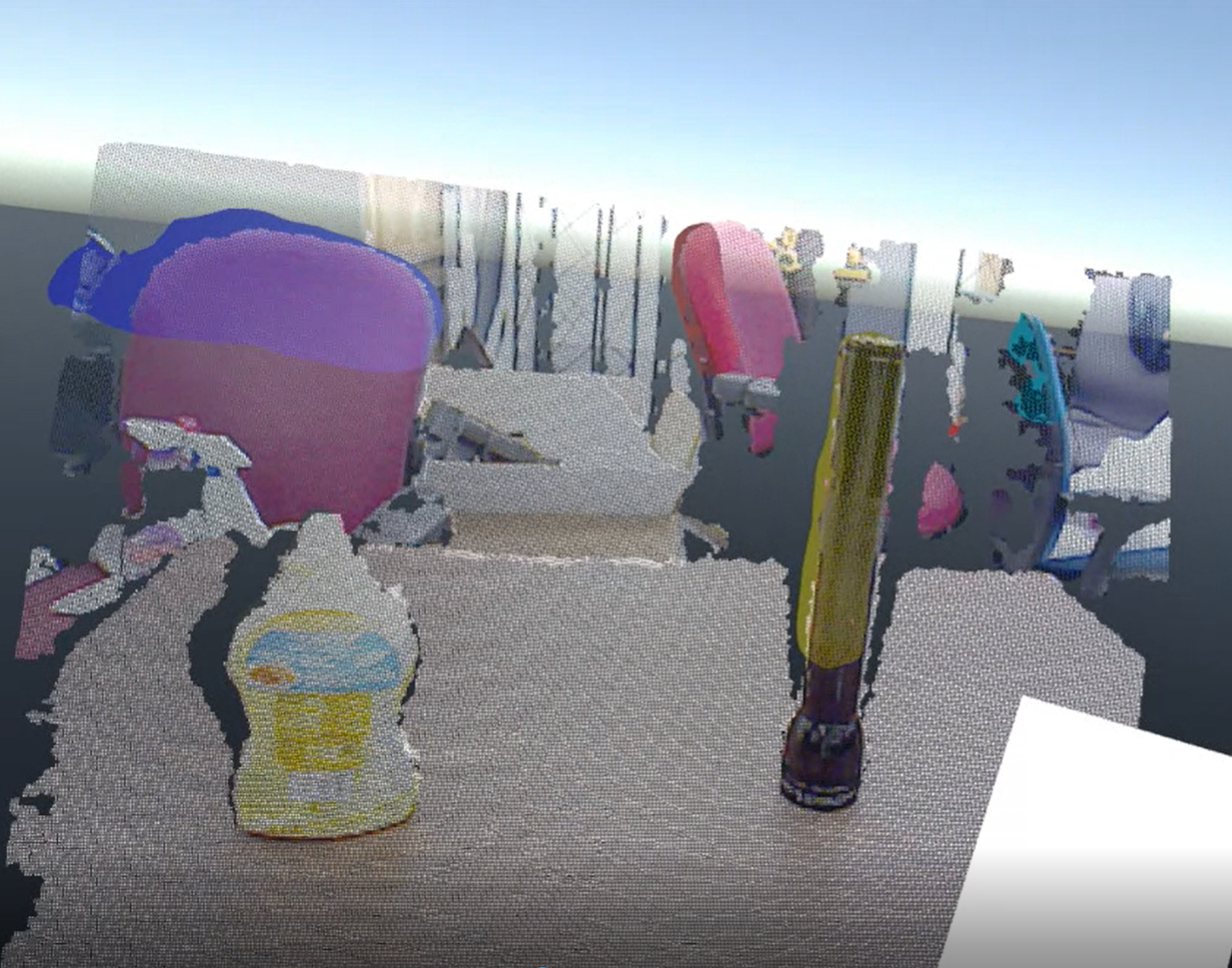}
    \caption{Scene Segmentation} 
    \label{segmentation} 
    \vspace{4ex}
  \end{minipage}%%
  \begin{minipage}[b]{0.5\linewidth}
    \centering
    \includegraphics[width=.9\linewidth]{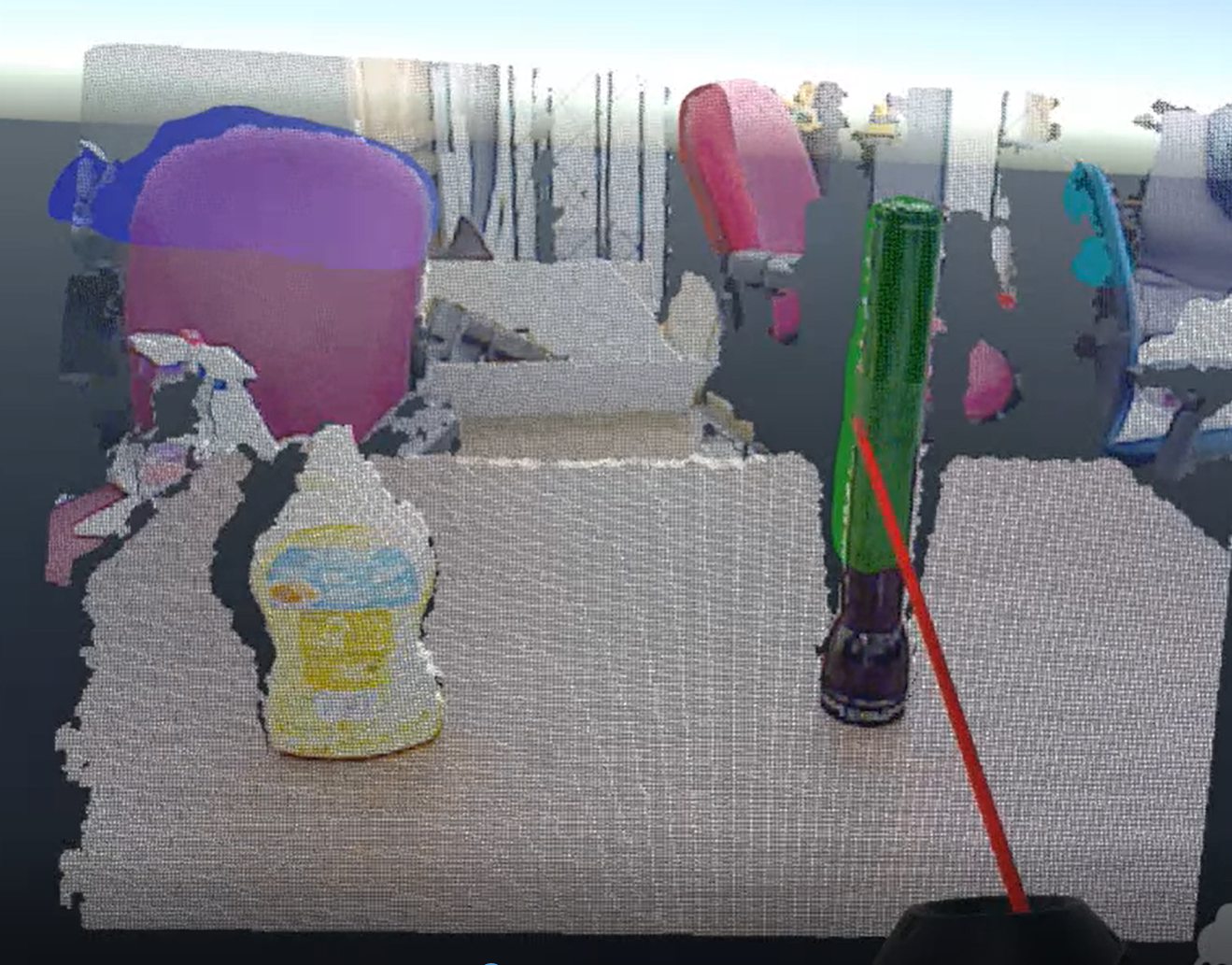}
    \caption{Object Selection} 
    \label{selection} 
    \vspace{4ex}
  \end{minipage} 
  \begin{minipage}[b]{0.5\linewidth}
    \centering
    \includegraphics[width=.9\linewidth]{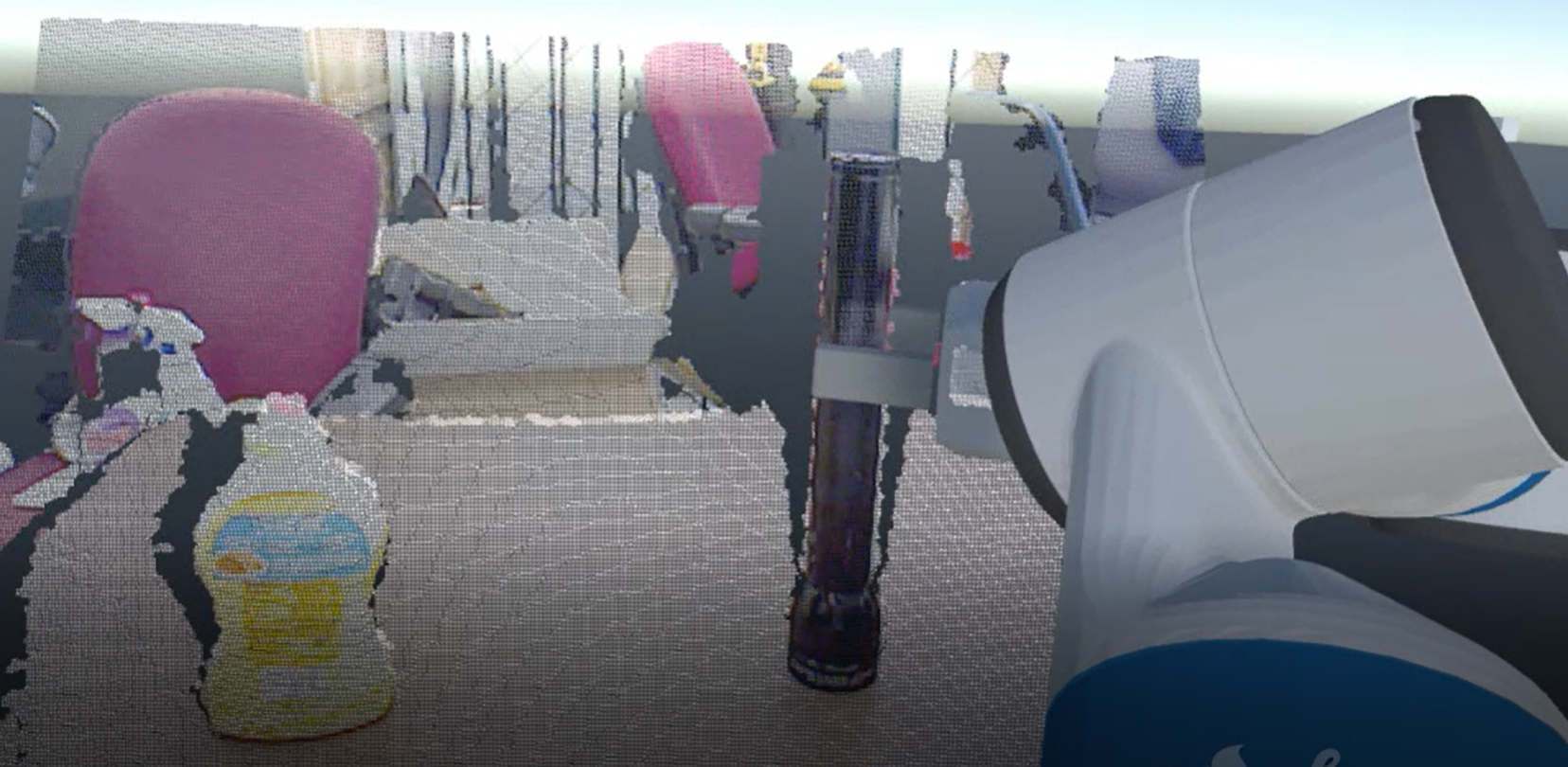}
    \caption{Grasping the Object} 
    \label{grasp} 
  \end{minipage}%% 
  \begin{minipage}[b]{0.5\linewidth}
    \centering
    \includegraphics[width=.9\linewidth]{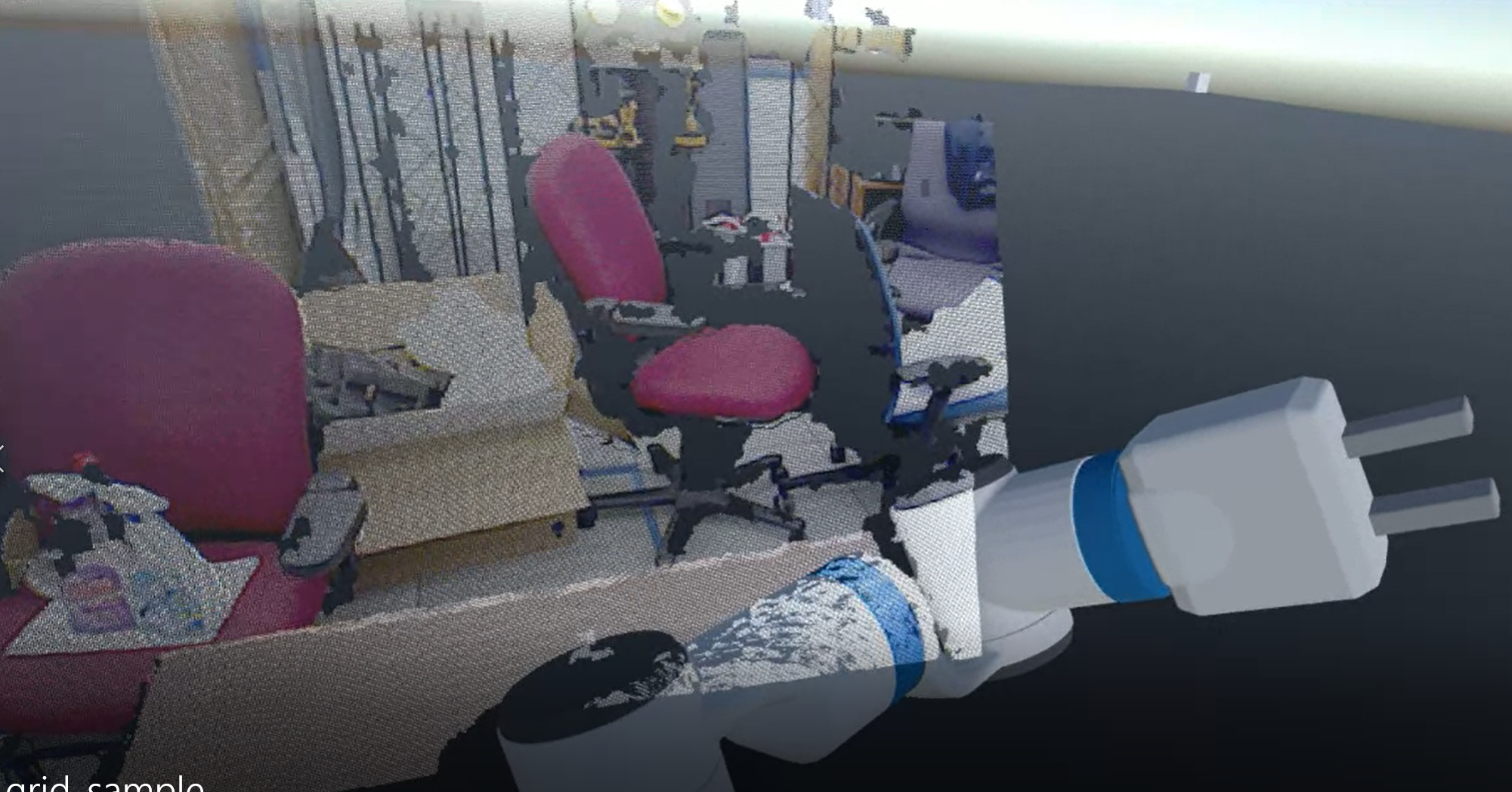}
    \caption{Lifting up the Object} 
    \label{lift} 
  \end{minipage} 
\end{figure}

\begin{algorithm}[h]
\SetAlgoLined
 \While{robot active}{
  Scan Scene\;
  Scene Segmentation\;
  \eIf{User Selected A Certain Object}{
   Complete the Mesh of the Given Object\;
   Plan Grasp\;
   Execute Grasp\;
   }
   {
   Continue\;
   }
 }
 \caption{Teleoperated Grasping}
 \label{alg:Teleoperated Grasping}
 
\end{algorithm}

The general workflow of our system can be summarized by Algorithm~\ref{alg:Teleoperated Grasping}. During each controlling cycle, the robot takes a pointcloud scan of the environment and produces a segmented pointcloud of objects detected in the pointcloud (Fig~\ref{segmentation}). The system then projects the segmented poincloud into the virtual reality space and waits for the user to select a certain mesh in the scene by pointing a virtual laser at it (Fig~\ref{selection}). Next, the partial pointcloud of the selected is shape-completed into a 3D mesh, and used to generate suitable grasps. A grasp is selected and the robot executes the grasp (Fig~\ref{grasp}) and picks up the object (Fig~\ref{lift}).

\begin{figure}[t]
\centering
\includegraphics[width=0.9\columnwidth]{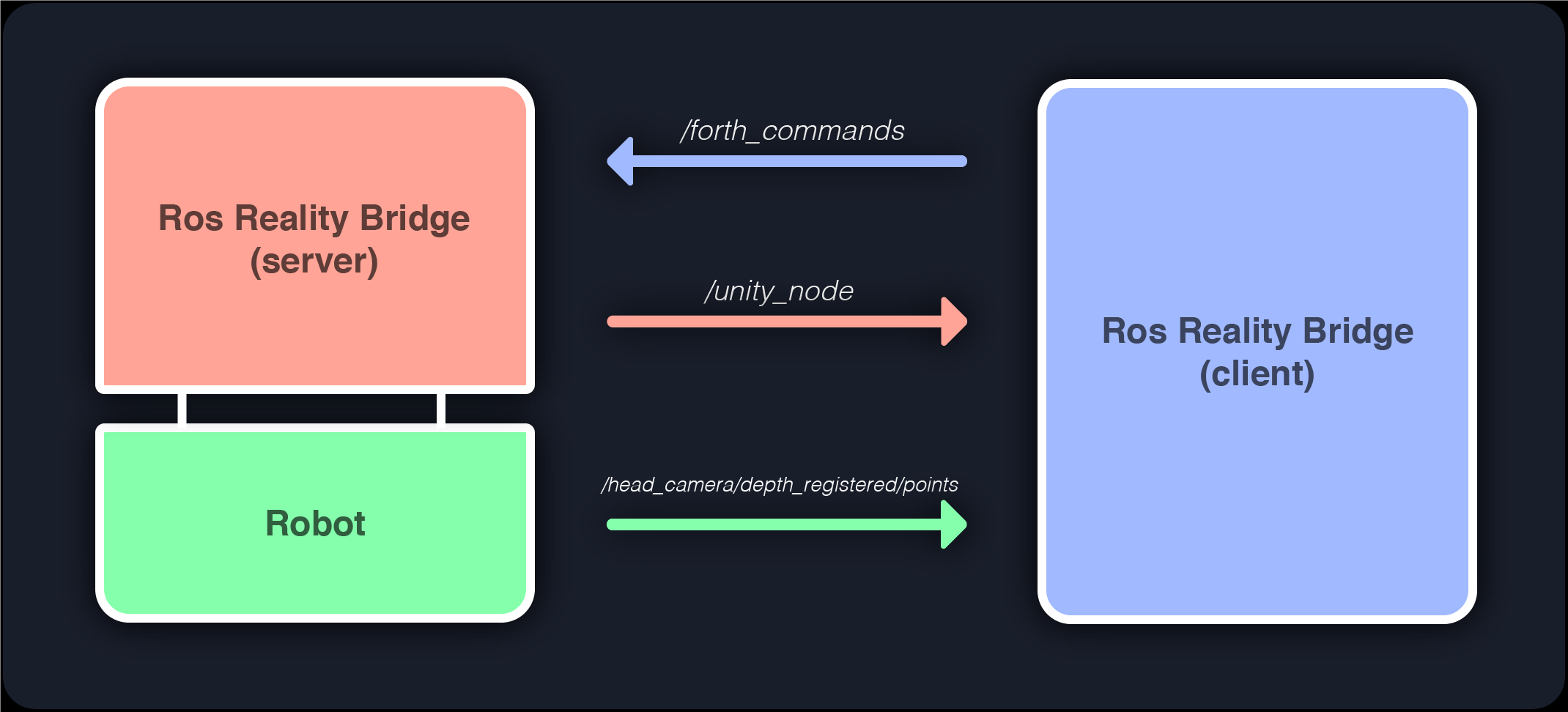} 
\caption{Modified Ros Reality Scheme}
\label{fig:scheme}
\end{figure}

\subsection{Virtual Reality Space}
The Virtual Reality portion of this project is an extension of Ros Reality\cite{rosreality}, through which users can visualize a robot and the surrounding scene using a HTC Vive headset. Two additional hand controllers are used to send commands to the robot. The setup is shown in Fig.~\ref{fig:robot_and_user}, with the user wearing the headset and holding the controllers. In order for the the system to be easily transferable to different robots, we modified Ros Reality to be more modular, shown in Fig.~\ref{fig:scheme}. Point clouds from the robot’s depth camera are broadcast and rendered as
meshes in the scene as well to allow the user to see what the robot sees. Using the Vive controllers, a
user can control all aspects of the robot. More specifically, controller input is sent over the ROS topic
/forth\_commands, which the server interprets to command the robot.

To improve pointcloud transmission speed and thus reduce delay, we made further modifications to the original system including: 1. Changing pointcloud encoding to BSON encoding; 2. Limiting the pointcloud update frequency; 3. Changing the web socket protocol to .NET protocol; 4. Setting up direct connection between robot and the client. With these modications, we were able to reduce pointcloud delay from 10 seconds to less than 1 second.

\subsection{Scene Segmentation}
Our scene segmentation algorithm is based on \cite{xiang2017darnn}. Our program takes in an RGB-D pointcloud and returns a semantic segmentation of the input pointcloud. We further utilized domain randomization \cite{tobin2017domain} to reduce noise in the generated semantic label.

\subsection{Shape Completion}
The shape completion program we used is a follow-up from our previous work \cite{varley2016shape}\cite{watkinsvalls2018multimodal}. We trained a convolution neural network (CNN) to predict the 3D mesh of an object given its partial pointcloud obtained from scene segmentation. A CNN is preferred over methods such as Gaussian Process Implicit Surface (GPIS) Completion \cite{GPIS} due to its faster run time, which is very important in our teleoperation loop. To further improve run speed, we set up a back-end shape completion server running over ros service so the front end doesn't need to run with GPU.

\subsection{Grasp Planning}
We used GraspIt \cite{graspit} for grasp planning. The program takes in a completed mesh and returns a proper grasp. Due to time concerns, we used grid sampling instead of simulated annealing for grasp sampling, so that we are only sampling around the principal axis of the given mesh. This allows us to increase sampling efficiency by an order of magnitude of 10 comparing to simulated annealing and correspondingly reduce run time from around 50 seconds to less than 3 seconds.

\section{CONCLUSIONS}
We developed a system that is capable of grasping and navigation tasks through teleoperation and can be seamlessly transferred to additional robotic agents. It is easy and intuitive to use, and can be further extended to serve as a powerful tool for collecting data under a user monitored fashion, that could be useful in fields such as computer vision and robot learning.

\addtolength{\textheight}{-12cm}   % This command serves to balance the column lengths

%%%%%%%%%%%%%%%%%%%%%%%%%%%%%%%%%%%%%%%%%%%%%%%%%%%%%%%%%%%%%%%%%%%%%%%%%%%%%%%%

%%%%%%%%%%%%%%%%%%%%%%%%%%%%%%%%%%%%%%%%%%%%%%%%%%%%%%%%%%%%%%%%%%%%%%%%%%%%%%%%

\bibliographystyle{IEEEtran}
\bibliography{IEEEabrv}

\end{document}